\pdfoutput=1

\PassOptionsToPackage{table}{xcolor} 
\documentclass[11pt]{article}

\usepackage[final]{acl}

\usepackage{times}
\usepackage{latexsym}

\usepackage[T1]{fontenc}

\usepackage[utf8]{inputenc}

\usepackage{microtype}

\usepackage{inconsolata}

\usepackage{graphicx}

\usepackage{amsmath}
\usepackage{amssymb}
\usepackage{bbm}
\usepackage{booktabs} 
\usepackage{fontawesome} 
\usepackage{makecell} 

\title{MoA: Heterogeneous Mixture of Adapters for Parameter-Efficient Fine-Tuning of Large Language Models}

\author{
 \textbf{Jie Cao\textsuperscript{1}},
 \textbf{Tianwei Lin \textsuperscript{1}},
 \textbf{Bo Yuan \textsuperscript{1}},
 \textbf{Rolan Yan \textsuperscript{2}},
 \textbf{Hongyang He},
 \\
 \textbf{Wenqiao Zhang\Thanks{Corresponding author} \textsuperscript{1}},
 \textbf{Juncheng Li \textsuperscript{1}},
 \textbf{Dongping Zhang },
 \textbf{Siliang Tang \textsuperscript{1}},
 \textbf{Yueting Zhuang \textsuperscript{1}}
\\
\\
 \textsuperscript{1} Zhejiang University,
 \textsuperscript{2} Tencent
\\
 \small{
   \{caojie, lintw, byuan, wenqiaozhang, junchengli, siliang, yzhuang\}@zju.edu.cn, rolanyan@qq.com
 }
}

\begin{document}
\maketitle

\begin{abstract}

Recent studies integrate Low-Rank Adaptation (LoRA) and Mixture-of-Experts (MoE) to further enhance the performance of parameter-efficient fine-tuning (PEFT) methods in Large Language Model (LLM) applications. 
Existing methods employ \emph{homogeneous} MoE-LoRA architectures composed of LoRA experts with either similar or identical structures and capacities. 
However, these approaches often suffer from representation collapse and expert load imbalance, which negatively impact the potential of LLMs.
To address these challenges, we propose a \emph{heterogeneous} \textbf{Mixture-of-Adapters (MoA)} approach. This method dynamically integrates PEFT adapter experts with diverse structures, leveraging their complementary representational capabilities to foster expert specialization, thereby enhancing the effective transfer of pre-trained knowledge to downstream tasks.
MoA supports two variants:
\textbf{(i)} \textit{Soft MoA} achieves fine-grained integration by performing a weighted fusion of all expert outputs;
\textbf{(ii)} \textit{Sparse MoA} activates adapter experts sparsely based on their contribution, achieving this with negligible performance degradation.
Experimental results demonstrate that heterogeneous MoA outperforms homogeneous MoE-LoRA methods in both performance and parameter efficiency.
Our project is available at \url{https://github.com/DCDmllm/MoA}.

\end{abstract}

\section{Introduction}
The rapid development of Large Language Models (LLMs)~\citep{achiam2023gpt,yang2025qwen3,grattafiori2024llama} is reshaping the field of NLP, demonstrating cross-domain generalization capabilities across a wide range of tasks. However, as model sizes continue to grow, the computational, storage, and deployment costs of traditional full fine-tuning methods increasingly become bottlenecks in practical applications. As a result, parameter-efficient fine-tuning (PEFT) methods \citep{houlsby_adapter_2019, he_towards_2022_paralleladapter, li_prefix-tuning_2021, lester_power_2021_prompttuning, zhang2024hyperllava, han2024parameter, lin2025healthgpt} are emerging as a key research direction for adapting LLMs. These methods introduce a small number of trainable, lightweight adapter modules on top of frozen pre-trained weights. By doing so, they effectively guide the activation and adjustment of the model’s existing knowledge representations while maintaining low overhead, enabling fast transfer and adaptation to specific downstream tasks.




The Low-Rank Adaptation (LoRA) method \citep{hu_lora_2021} reparameterizes weight matrices with low-rank decomposition, enabling the approximation of full fine-tuning using only a small number of trainable parameters. However, increasing the rank does not consistently improve performance, as the representational capacity of LoRA tends to saturate \citep{chen2022revisiting, zhu_sira_2023}. To overcome this limitation, recent methods integrate LoRA into the Mixture-of-Experts (MoE) architecture \citep{shazeer2017outrageously_sparsemoe}, giving rise to the \textit{MoE-LoRA} framework \citep{zadouri_pushing_2023, zhu_sira_2023}. In this setup, multiple structurally identical LoRA experts are dynamically routed at the token level via a learnable routing mechanism, thereby enhancing the model’s adaptability to diverse inputs and increasing its representational flexibility.

\begin{figure*}[ht]
  \includegraphics[width=1\textwidth]{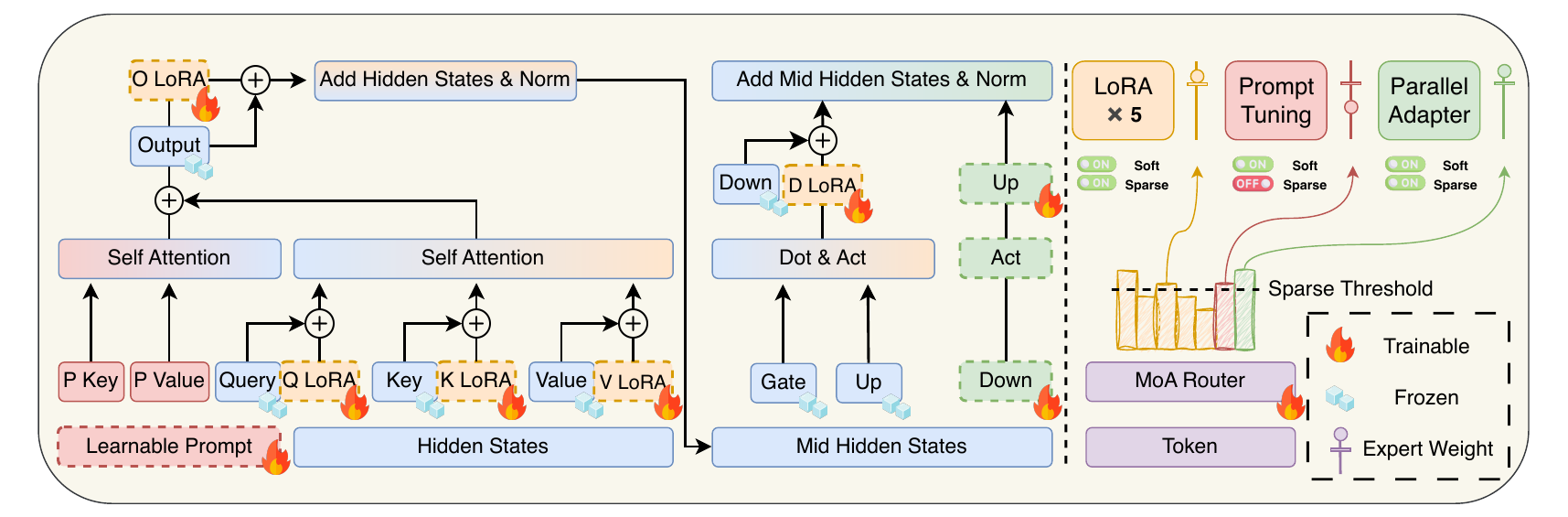}
  \caption{MoA architecture with heterogeneous PEFT adapters. It is worth noting that in Sparse MoA, the Prompt Tuning module is deactivated due to its non-token-level activation mechanism.}
  \label{fig:moa}
\end{figure*}



Although MoE-LoRA hybrids enhance representational capacity by introducing multiple LoRA experts, their homogeneous design causes the experts to tend toward learning similar representations during training, leading to \textbf{representation collapse} \citep{chi2022representation, wang_hmoe_2024, lin2024teamlora}. This undermines expert diversity and specialization. Moreover, the dynamic routing mechanism is prone to \textbf{expert load imbalance} in homogeneous architectures, where a few initially well-performing experts consistently receive the majority of tokens, suppressing the participation of other experts. These mechanisms ultimately result in redundant experts \citep{zhou2022mixture}, which limit the expressiveness of the model and reduce the efficiency of resource utilization.

Based on the above analysis, we posit that the representational convergence of homogeneous experts may limit the performance ceiling of PEFT methods. To address the issues of redundant experts and load imbalance in MoE-LoRA, we propose a \textbf{Mixture-of-Adapters (MoA)} approach with heterogeneous structures (Figure~\ref{fig:moa}). MoA consists of PEFT adapter experts with diverse architectural forms and employs token-level dynamic routing to activate experts on demand. By leveraging the complementary representational capacities of different structures, MoA promotes expert specialization through functional differentiation among experts. This mechanism enables more refined utilization of pre-trained knowledge and enhances the ability of models to transfer and generalize across diverse downstream tasks. Our main contributions are as follows:

\textbf{(i)} We propose MoA based on heterogeneous experts, which constructs adapters with complementary representational capacities by integrating structurally diverse PEFT modules. This effectively enhances expert specialization, mitigating the issue of expert redundancy commonly observed in traditional MoE-LoRA architectures. In addition, MoA achieves efficient adaptation to downstream tasks using fewer trainable parameters, fully leveraging the knowledge representations embedded in the pre-trained model.

\textbf{(ii)} Building on this foundation, we further design two variants: \textbf{Soft MoA} and \textbf{Sparse MoA}. In Sparse MoA, a threshold function dynamically selects active experts that valuable contribute to the current input. This reduces redundant computation while enabling the model to engage more experts for critical tokens, thereby improving representational capacity.

\textbf{(iii)} We systematically evaluate the architectural advantages and practical effectiveness of MoA across multiple tasks. Experimental results show that MoA significantly outperforms SoTA MoE-LoRA methods in GPU memory usage, training efficiency, and inference speed. Moreover, it achieves \textbf{higher parameter efficiency and knowledge transfer capability while maintaining or even improving downstream performance.}

\section{Related Works}








\subsection{Parameter-Efficient MoE}
In non-PEFT scenarios, standard sparse MoE \citep{shazeer2017outrageously_sparsemoe, fedus_switch_2022, jiang2024mixtral} architectures must employ discrete top-k routing strategies to reduce computational overhead. In contrast, under PEFT settings, where PEFT experts occupy significantly less parameters than the pre-trained LLMs, it becomes feasible to compute a soft-weighted output by averaging experts’ predictions according to the router’s distribution. Thus, we categorize PEFT-based MoE into two paradigms: \textbf{sparse parameter efficient MoE} and \textbf{soft-weighted parameter efficient MoE}.

\paragraph{Sparse parameter efficient MoE}
Building on the demonstrated superiority of standard sparse MoE architectures, recent works has integrated sparse MoE with parameter-efficient fine-tuning (PEFT) methods to enhance resource-constrained adaptation performance. Most of these approaches are based on LoRA, leveraging its superior efficacy in parameter-efficient adaptation as a foundational component. SiRA \citep{zhu_sira_2023} enforces top-$k$ experts routing with per-expert capacity constraints, limiting the maximum tokens each expert processes, and uses dropout to the gating network to mitigate overfitting. MoELoRA \cite{luo_moelora_2024} employs contrastive learning to encourage experts to learn different features, thus mitigating the random routing phenomenon observed in MoE. To reduce redundancy, MoLA \citep{gao_higher_2024} explores layer-wise expert redundancy by assigning a different number of LoRA experts to each Transformer layer, revealing that the lower layers exhibit higher redundancy in the sparse MoE-LoRA frameworks. Advancing this line, AdaMoLE \citep{liu_adamole_2024} dynamically adjusts the activation of LoRA experts for each input token via a threshold network integrated into the top-$k$ routing.

\paragraph{Soft-weighted parameter efficient MoE}
 MoLoRA \cite{zadouri_pushing_2023} proposes a soft-weighted MoE-LoRA method with a token-level routing strategy. LoRAMoE \cite{dou_loramoe_2024} integrates LoRA experts using a linear router to alleviate the forgetting of the world knowledge previously stored in LLMs. MOLE \cite{wu_mixture_2024} proposes to efficiently compose multiple trained LoRAs while preserving all their characteristics by training only the router within each layer. HydraLoRA \cite{tian_hydralora_2024} introduces an asymmetric structure with a shared matrix, further enhancing parameter efficiency compared to other MoE-LoRA methods. 
Diverging from methods that ensemble the outputs of individual experts, certain approaches such as AdaMix \cite{wang_adamix_2022}, SMEAR \cite{muqeeth_soft_2024}, and MoV \citep{zadouri_pushing_2023} adopt a strategy of first merging expert parameters via routing, followed by computation using the merged parameters. This aims to reduce computational overhead. 


\subsection{Heterogeneous MoE}
AutoMoE \cite{jawahar_automoe_2023} employs the neural architecture search to obtain heterogeneous MoE subnetworks from original FFN experts on a small-scale MoE model that utilizes the top-1 routing strategy. HMoE \cite{wang_hmoe_2024} constructs heterogeneous experts by varying the dimensions of FFN experts, thereby endowing them with diverse capacities. UNIPEFT \cite{mao_unipelt_2022} employs distinct gated networks for various PEFT methods in instance-level to adapt to different examples. 

\section{Method}





Homogeneous MoE methods inherently suffer from the representation collapse and load imbalance issues, fundamentally stemming from their experts' identical structures and capacities, which lead to insufficient specialization of individual experts. To address this, we construct inherently heterogeneous MoE experts leveraging structurally diverse and positionally varied PEFT adapters for each Transformer block.

LoRA, Parallel Adapter, and Prompt Tuning approaches can be simplified into a unified view \citep{he_towards_2022_paralleladapter}:
\begin{equation}
    \boldsymbol{h} = F(\boldsymbol{x}) +  E(\boldsymbol{x}),
\end{equation}
where $F(\boldsymbol{x})$ denotes the transformer module to which the corresponding PEFT Adapter \(E\) is applied. Our selection of heterogeneous PEFT experts comprises LoRA \citep{hu_lora_2021}, Parallel Adapters \citep{he_towards_2022_paralleladapter}, and the zero-initialized Prompt Tuning \citep{zhang_llama-adapter_2023}, the details of these methods and the unified view are shown in Appendix~\ref{sec:peft methods}.

\subsection{Soft MoA}

In each transformer layer, our Soft MoA comprises a soft-weighted router $R$ and a set of experts $\{E_1, E_2, \dots, E_n\}$ that are composed of PEFT adapters of different types and placements.
Our router network commonly consists of a dense layer with trainable weights $\boldsymbol{W}_r \in \mathbb{R}^{d \times n}$ followed by a sigmoid function which takes an intermediate token representation $\boldsymbol{x}$ as input and output weights $R(\boldsymbol{x})$ for experts: 
\begin{equation}
    R(\boldsymbol{x}) = {\rm Sigmoid}(\boldsymbol{W}_{r}^{T}\boldsymbol{x}).
\end{equation}
Then, each expert's weight \(R(\boldsymbol{x})_i\) is applied to its output \(E_{i}(\boldsymbol{x})\), controlling the influence of expert \(E_i\) in that transformer layer:
\begin{equation}
    \boldsymbol{h} = F_{i}(\boldsymbol{x}) +  R(\boldsymbol{x})_{i} E_{i}(\boldsymbol{x}).
\end{equation}

Unlike previous MoE methods \cite{zadouri_pushing_2023, dou_loramoe_2024}, our MoA leverages heterogeneous experts of various types within the Transformer module to process each token. Due to the diverse characteristics and capacities of these heterogeneous experts, MoA neither suffer from the representation collapse problem nor need a load-balancing loss. Moreover, MoA employs a sigmoid function in the router, rather than the traditional softmax activation used in conventional MoE \cite{shazeer2017outrageously_sparsemoe, fedus_switch_2022}, encouraging cooperation among experts instead of competition and fully capitalizing on their unique specializations.

\subsection{Sparse MoA}
Methods based on soft-weighted routing, including MoA and other MoE approaches \cite{zadouri_pushing_2023, dou_loramoe_2024, muqeeth_soft_2024}, require computing experts \(E_i\) even when their weights \(R(\boldsymbol{x})_i \approx 0\), despite negligible contributions to outputs, resulting in unnecessary computational overhead and time consumption. To address this issue, we propose Sparse MoA, which dynamically computes only the necessary experts, thereby avoiding the computation of low-contributing ones.

To implement sparse MoA, a straightforward approach is to apply a fixed threshold $\Gamma$ to the experts' weights. An expert $E_i$ is activated for computation if and only if its weight satisfies $R(\boldsymbol{x})_i > \Gamma$, where unselected experts are entirely excluded from the computational graph, thus eliminating redundant computations. The detailed computation is defined as follows:  
\begin{equation}
    \boldsymbol{h} = 
        \begin{cases}
        F_{i}(\boldsymbol{x}) + R(\boldsymbol{x})_{i} E_{i}(\boldsymbol{x}) & \text{if } R(\boldsymbol{x})_{i} > \Gamma, \\
        F_{i}(\boldsymbol{x}) & \text{otherwise}.
        \end{cases}
\end{equation}
Note that when the expert $E_i$ is selected, its weight $R(\boldsymbol{x})_i$ is still used to control its influence. The threshold-based approach can activate varying numbers of heterogeneous experts for each token, dynamically adapting to the current task and data. 

However, due to the varying semantic importance of tokens in contexts \cite{xia2025tokenskip}, assigning token-specific thresholds can more effectively harness multi-expert collaboration: lower thresholds for critical tokens promote broader expert engagement, while higher thresholds for less critical tokens restrict processing to fewer specialized experts. Inspired by AdaMoLE \citep{liu_adamole_2024}, we replace fixed thresholds with a learnable threshold function. Specifically, the threshold for each token is calculated via a linear layer followed by a sigmoid activation: 
\begin{equation}
    \Gamma = \Gamma_{max}{\rm Sigmoid}(\boldsymbol{W}_{\Gamma}^T \boldsymbol{x} + \boldsymbol{b}_{\Gamma}),
\end{equation}
where $\boldsymbol{W}_{\Gamma}^T \in \mathbb{R}^{d \times 1}$ and $\boldsymbol{b}_{\Gamma} \in \mathbb{R}^{1}$ are trainable parameters, $\Gamma_{max}$ is a hyperparameter that defines the upper bound of threshold value.

As the core objective of PEFT methods is to fully leverage LLMs' pre-trained knowledge and capabilities via adapters for downstream tasks \citep{ghosh_closer_2024}, Sparse MoA enhances this efficiency by adaptively activating diverse combinations of heterogeneous experts. Moreover, Sparse MoA adjusts the number of participating experts based on the contextual importance of tokens, thereby mitigating redundant expert computations. In extreme cases where certain tokens require no expert intervention and can be processed solely by the frozen LLM, Sparse MoA deactivates all experts, thus maximizing computational efficiency.

Notably, due to its computational mechanism, Prompt Tuning does not facilitate token-level expert allocation within a batch. Consequently, Prompt Tuning experts are not included in our Sparse MoA framework.
\section{Experiments}
\subsection{Datasets}
We conduct extensive experiments and analyses on mathematical and commonsense reasoning tasks. Details for code generation tasks are provided in Appendix~\ref{sec:code_task}. For mathematical reasoning, we evaluate performance on six test datasets: GSM8K \cite{cobbe2021gsm8k}, SVAMP \cite{patel2021SVAMP}, MultiArith \cite{roy2016MultiArith}, AddSub \cite{hosseini2014AddSub}, AQuA \cite{ling2017AQuA}, and SingleEq \cite{koncel2015SingleEQ}. The training dataset is Math14K constructed by \cite{hu_llm-adapters_2023}, which includes the training sets of GSM8K and AQuA, augmented with rationales generated by ChatGPT and GPT-4. For commonsense reasoning, we test on eight datasets: BoolQ \cite{clark2019boolq}, PIQA \cite{bisk2020piqa}, SIQA \cite{sap2019socialiqa}, HellaSwag, WinoGrande \cite{sakaguchi2021winogrande}, ARC Challenge (ARC-c), ARC Easy (ARC-e) \cite{chollet2019ARC}, and OBQA \cite{mihaylov2018OBQA}. The training dataset for commonsense reasoning, Commonsense15K, is sampled from Commonsense170K \cite{hu_llm-adapters_2023} which comprises eight training sets.

\begingroup
\setlength{\tabcolsep}{4pt}
\begin{table*}[ht]
\small
  \centering
  \begin{tabular}{lcccccccc}
    \Xhline{1.5pt}
    \rowcolor[HTML]{DAE0FB}
    \textbf{Models} & \textbf{AddSub} & \textbf{AQuA} & \textbf{GSM8k} & \textbf{MultiArith} & \textbf{SingleEq} & \textbf{SVAMP} & \textbf{Average} & \textbf{Param} \\
    \hline
    \hline
    \rowcolor[HTML]{D3D3D3}
    LoRA & $87.68_{\pm 1.14}$ & $36.75_{\pm 0.99}$ & $76.14_{\pm 0.88}$ & $96.28_{\pm 0.79}$ & $96.98_{\pm 0.30}$ & $81.27_{\pm 0.45}$ & $79.18_{\pm 0.26}$ & 23.06M \\ 
    MoLoRA & $91.73_{\pm 1.02}$ & $38.06_{\pm 1.86}$ & $77.46_{\pm 0.68}$ & $\boldsymbol{99.33}_{\pm 0.17}$ & $97.18_{\pm 0.41}$ & $81.77_{\pm 0.68}$ & $80.92_{\pm 0.19}$ &  100.14M \\
    HydraLoRA & $91.22_{\pm 0.15}$ & $39.11_{\pm 2.17}$ & $77.51_{\pm 0.59}$ & $99.17_{\pm 0.33}$ & $96.85_{\pm 0.34}$ & $81.87_{\pm 0.25}$ & $80.95_{\pm 0.28}$ & 45.09M\\
    MoLA & $90.46_{\pm 0.53}$ & $\underline{39.37}_{\pm 1.80}$ & $77.61_{\pm 0.61}$ & $98.94_{\pm 0.19}$ & $97.05_{\pm 0.20}$ & $82.27_{\pm 0.60}$ & $80.95_{\pm 0.14}$ & 100.14M \\ 
    AdaMoLE & $90.97_{\pm 0.53}$ & $\boldsymbol{40.03}_{\pm 2.37}$ & $77.51_{\pm 0.27}$ & $98.33_{\pm 2.05}$ & $97.31_{\pm 0.30}$ & $82.40_{\pm 0.50}$ & $81.09_{\pm 0.78}$ & 101.12M \\ 
    FlyLoRA & $89.45_{\pm 1.83}$ & $34.65_{\pm 0.68}$ & $77.46_{\pm 1.10}$ & $97.11_{\pm 0.67}$ & $96.52_{\pm 0.93}$ & $82.87_{\pm 0.64}$ & $79.68_{\pm 0.63}$ & 23.06M \\
    \hline
    \rowcolor[HTML]{E9F3FE}
    Soft MoA & $\boldsymbol{92.24}_{\pm 0.89}$ & $38.98_{\pm 1.57}$ & $\underline{78.01}_{\pm 0.23}$ & $98.94_{\pm 0.54}$ & $\boldsymbol{97.44}_{\pm 0.71}$ & $\boldsymbol{83.43}_{\pm 0.55}$ & $\boldsymbol{81.51}_{\pm 0.35}$ & 24.52M\\ 
    \rowcolor[HTML]{E9F3FE}
    Sparse MoA & $\underline{91.90}_{\pm 0.44}$ & $38.06_{\pm 2.27}$ & $\boldsymbol{78.09}_{\pm 0.95}$ & $\underline{99.17}_{\pm 0.17}$ & $\underline{97.31}_{\pm 0.23}$ & $\underline{82.70}_{\pm 0.00}$ & $\underline{81.20}_{\pm 0.47}$ &  22.29M\\ 
   \Xhline{1.5pt}
  \end{tabular}
  \caption{\label{table:math14k}
  Experiment results on mathematical reasoning benchmarks. The evaluation metric is accuracy. All methods are run with three random seeds; mean and standard deviation are reported. ``Param'' indicates trainable parameters.
  }
\end{table*}

\begingroup
\setlength{\tabcolsep}{0pt}
\begin{table*}[ht]
\small
  \centering
  \begin{tabular}{lcccccccccc}
    \Xhline{1.5pt}
    \rowcolor[HTML]{DAE0FB}
    \textbf{Models} & \textbf{BoolQ} & \textbf{PIQA} & \textbf{SIQA} & \textbf{HellaSwag} & \textbf{WinoGrande} & \textbf{ARC-C} & \textbf{ARC-E} & \textbf{OBQA} & \textbf{Average} & \textbf{Param} \\
    \hline
    \hline
    \rowcolor[HTML]{D3D3D3}
    LoRA& $72.94_{\pm 0.29}$  & $86.22_{\pm 0.64}$  & $78.78_{\pm 0.64}$  & $86.34_{\pm 1.37}$  & $80.01_{\pm 1.87}$  & $82.08_{\pm 0.51}$  & $92.63_{\pm 0.98}$  & $86.93_{\pm 0.81}$  & $83.24_{\pm 0.58}$ & 23.06M \\ 
    MoLoRA & $\boldsymbol{73.89}_{\pm 0.77}$  & $87.32_{\pm 0.73}$  & $78.98_{\pm 0.96}$  & $\underline{88.52}_{\pm 0.58}$  & $82.11_{\pm 0.41}$  & $84.84_{\pm 1.04}$  & $93.45_{\pm 0.45}$  & $87.33_{\pm 0.99}$  & $84.56_{\pm 0.58}$ & 100.14M \\
    HydraLoRA & $73.52_{\pm 0.25}$  & $86.65_{\pm 0.65}$  & $79.17_{\pm 0.42}$  & $87.46_{\pm 1.00}$  & $82.30_{\pm 0.39}$  & $83.87_{\pm 0.45}$  & $93.28_{\pm 0.49}$  & $86.73_{\pm 1.50}$  & $84.09_{\pm 0.31}$ & 45.09M \\
    MoLA & $73.19_{\pm 0.42}$  & $87.01_{\pm 0.44}$  & $79.32_{\pm 0.75}$  & $87.60_{\pm 0.67}$  & $\underline{82.53}_{\pm 0.90}$  & $84.95_{\pm 0.42}$  & $93.35_{\pm 0.33}$  & $87.60_{\pm 0.92}$  & $84.45_{\pm 0.22}$ & 100.14M \\
    AdaMoLE &  $73.60_{\pm 0.37}$  & $\underline{88.08}_{\pm 0.61}$  & $\boldsymbol{79.72}_{\pm 0.48}$  & $\boldsymbol{88.70}_{\pm 0.65}$  & $81.29_{\pm 1.14}$  & $85.04_{\pm 0.81}$  & $93.27_{\pm 0.19}$  & $\underline{89.13}_{\pm 0.81}$  & $\underline{84.85}_{\pm 0.21}$ & 101.12M \\ 
    \hline
    \rowcolor[HTML]{E9F3FE}
    Soft MoA & $72.86_{\pm 0.40}$ & $\boldsymbol{88.08}_{\pm 0.55}$ & $\underline{79.32}_{\pm 0.72}$ & $87.82_{\pm 0.44}$ & $\boldsymbol{83.03}_{\pm 0.70}$ & $\boldsymbol{85.84}_{\pm 0.31}$ & $\underline{93.49}_{\pm 0.46}$ & $\boldsymbol{89.27}_{\pm 0.50}$ & $\boldsymbol{84.96}_{\pm 0.26}$ & 24.52M \\  
    \rowcolor[HTML]{E9F3FE}
    Sparse MoA & $\underline{73.69}_{\pm 0.31}$ & $87.36_{\pm 0.27}$ & $78.51_{\pm 0.23}$ & $87.97_{\pm 0.24}$ & $81.66_{\pm 0.91}$ & $\underline{85.55}_{\pm 0.13}$ & $\boldsymbol{93.53}_{\pm 0.09}$ & $88.67_{\pm 0.58}$ & $84.62_{\pm 0.09}$ & 22.29M\\ 
   \Xhline{1.5pt}
  \end{tabular}
  \caption{\label{table:commonsense15k}
  Experiment results on commonsense reasoning benchmarks. The evaluation metric is accuracy.
  }
  \vspace{-2mm}
\end{table*}

\endgroup

\subsection{Baseline Models}


To validate the efficacy of our method, we compare it against: (i) the standard LoRA\citep{hu_lora_2021}; (ii) SoTA soft-weighted parameter-efficient MoE methods: MoLoRA \cite{zadouri_pushing_2023} and HydraLoRA \cite{tian_hydralora_2024}; (iii) SoTA sparse parameter-efficient MoE approaches: MoLA \citep{gao_higher_2024} and AdaMoLE \cite{liu_adamole_2024}; (iv) other LoRA-based methods DenseLora \cite{mu2025denselora} and FlyLoRA \cite{zouflylora}.
\subsection{Experiment Settings} \label{sec:setting}
Our experiments use the LLaMA3.1-8B model \citep{grattafiori2024llama3}, Qwen3-8B model, and the larger Qwen3-14B model \cite{yang2025qwen3}. To ensure fair comparisons, we apply identical configurations to all LoRA-based methods: LoRA is inserted into four weight matrices in the self-attention module ($W_q$, $W_k$, $W_v$, $W_o$) and one in the FFN module ($W_{\text{down}}$). For our MoA methods and all MoE-LoRA baselines, we set the number of experts to 8, with both the rank and alpha fixed at 8. To ensure a fair comparison with our MoA methods, we adjust the ranks of LoRA, FlyLoRA, and DenseLoRA to 16, 32, and 256, respectively, to maintain an equivalent number of trainable parameters. MoA integrates 7 heterogeneous experts, including five LoRA modules, parallel adapters in the FFN layer, and a zero-initialized prompt-tuning module. The bottleneck dimension of the parallel adapter is set to 16, and the prompt length to 10. We use the AdamW optimizer with a learning rate of 6e-3. For commonsense reasoning, input sequences are truncated to 200 tokens; for mathematical reasoning, to 300 tokens. All models are trained on A6000-48G GPUs.

The experimental results on Qwen3-8B and the larger Qwen3-14B are given in the appendix~\ref{appendix:qwen3_results}.


\subsection{Experiment Results}

As shown in Tables~\ref{table:math14k} and~\ref{table:commonsense15k}, Soft MoA and Sparse MoA consistently outperform homogeneous MoE-LoRA baselines on both mathematical and commonsense reasoning tasks. Soft MoA achieves the highest accuracy on math benchmarks (81.51\%) with only 24.52M trainable parameters—nearly 4× fewer than AdaMoLE and MoLoRA. Sparse MoA delivers strong performance (81.20\% math, 84.62\% commonsense) with the smallest parameter count (22.29M), surpassing all other methods in mathematical accuracy. These results demonstrate that heterogeneous experts in \textbf{MoA enhance both performance and efficiency, effectively addressing redundancy in existing MoE-LoRA designs.}




\section{Ablation Study}

\begin{figure*}[t]
  \includegraphics[width=1\textwidth]{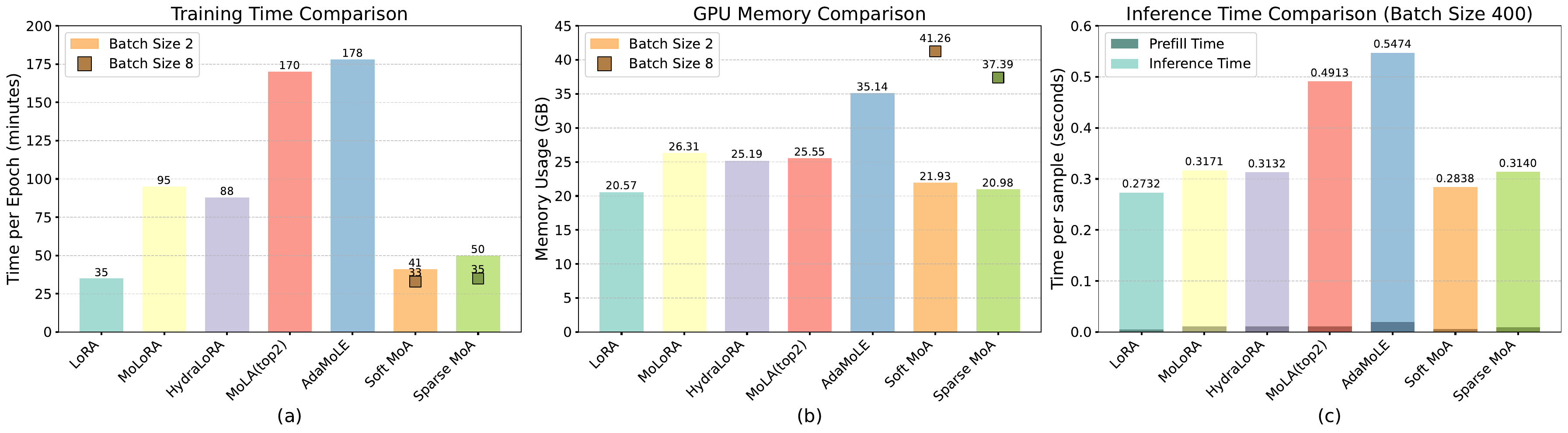}
  \caption{Comparison of different models under identical batch sizes (1* 48G GPU): (a) Training time per epoch, (b) GPU memory consumption during training, and (c) Average inference time per sample.}
  \label{fig:time_memoery}
\end{figure*}
\subsection{Ablation Study on Soft MoA Components}
\paragraph{The token-level soft-weighted router effectively leverages the distinct characteristics of heterogeneous experts.}
Table~\ref{table:ablation_softMoA} shows the performance of individual PEFT methods and their naive combination on two reasoning tasks. Among the three standalone PEFT methods, LoRA achieves the highest accuracy. Notably, the naive composition of these three PEFT methods degrades performance, failing to effectively leverage the functionalities of diverse experts. In contrast, our Soft MoA integrates seven heterogeneous experts via a token-level router, achieving significant improvements over standalone LoRA (+2.33 on mathematical reasoning, +1.72 on commonsense reasoning). This demonstrates that Soft MoA's router dynamically assigns experts based on token-specific contextual demands, fully exploiting the complementary strengths of these structurally diverse modules.

\begingroup
\begin{table}[t]
\small
  \centering
  \begin{tabular}{lccc}
    \Xhline{1.5pt}
    \rowcolor[HTML]{DAE0FB}
    \textbf{Modules} & \textbf{Router} & \textbf{Commonsense} & \textbf{Math}  \\ 
    \hline
    \hline
    Prompt-tuning & \faClose & 78.03 & 78.06 \\
    Parallel Adapter (PA) & \faClose &  81.62 & 79.20 \\
    LoRA & \faClose &  83.24  & 79.18 \\
    LoRA & Sigmoid R & \underline{84.23} & \underline{80.50} \\
    LoRA+Prompt+PA & \faClose & 81.58 & 78.28 \\
    LoRA+Prompt+PA & Softmax R & 84.16 & 80.21 \\
    \hline
    \rowcolor[HTML]{E9F3FE}
    Soft MoA & Sigmoid R & \textbf{84.96} &  \textbf{81.51} \\
    \Xhline{1.5pt}
  \end{tabular}
  \caption{\label{table:ablation_softMoA}
  Ablation study of each type of PEFT expert and the activation function of the Router in MoA. 
  }
\end{table}

\paragraph{The sigmoid activation function outperforms softmax in the MoA router.}
Conventional MoE methods with homogeneous experts typically use routers with softmax activation, which enforces weight trade-offs over experts ($\sum_i R(\boldsymbol{x})_i = 1$). For comparison, we replaced the sigmoid function with softmax in our framework. As shown in Table~\ref{table:ablation_softMoA}, while the softmax-based router improves upon standalone LoRA and the naive composition, it underperforms MoA equipped with a sigmoid router. This indicates a fundamental operational difference: homogeneous MoE experts function within a competitive paradigm, whereas MoA's heterogeneous experts leverage collaborative computation (sigmoid permits non-exclusive activation). The sigmoid router dynamically scales expert contributions - enabling full-capacity collaboration ($R(\boldsymbol{x})_i \approx 1$ for all experts) or complete deactivation ($R(\boldsymbol{x})_i \approx 0$)—thereby maximizing the utility of architectural diversity.
\paragraph{The five LoRA modules contribute the most, while the Parallel Adapter and Prompt Tuning further improve performance.}
The superior performance of LoRA over Prompt-tuning and Parallel Adapter, combined with its inherent inclusion of positionally diverse expert modules, motivated our extension of LoRA’s multi-module experts with a token-level router. While this adaptation yielded significant improvements over standalone LoRA (e.g., +1.32 on mathematical reasoning), it underperformed the full MoA (-1.01), highlighting two critical insights:
1. MoA’s primary strength derives from leveraging multiple LoRA modules as heterogeneous experts.
2. The distinct mechanisms of Prompt Tuning and Parallel Adapter further complement these LoRA-based experts.


\section{In-depth Analysis}

\begin{figure*}[t]
  \includegraphics[width=1\textwidth]{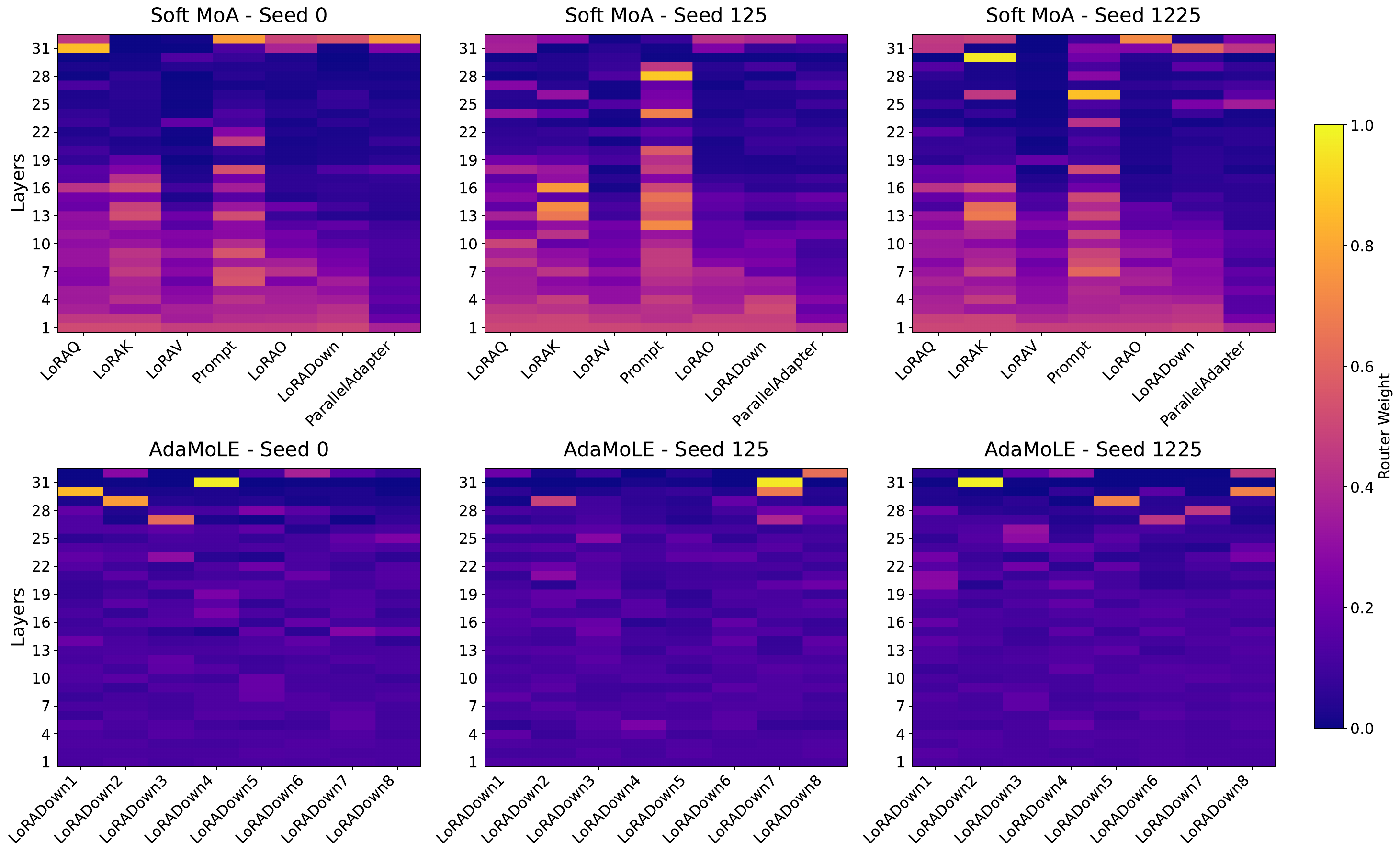}
  \caption{Comparison of router weight distributions between Soft MoA and AdaMoLE on the BoolQ test set under different random seeds. The MoA method exhibits strong consistency, whereas AdaMoLE does not.}
  \label{fig:seeds_heatmap}
\end{figure*}

\subsection{Efficiency Comparison}

To systematically compare the differences between Soft MoA, Sparse MoA, and baseline methods, Figure~\ref{fig:time_memoery} presents the training time per epoch, GPU memory requirements, and average inference latency per sample under identical configurations (single GPU setup, fixed batch size (BS)).

As shown in Figure~\ref{fig:time_memoery}(a), under identical batch size conditions (BS=2), both Soft MoA and Sparse MoA require significantly less training time compared to other MoE methods. Notably, Soft MoA achieves less than half the training time of MoLoRA and HydraLoRA. This efficiency advantage stems not only from reduced parameter counts but also from architectural optimizations - for instance, while HydraLoRA contains half the parameters of MoLoRA, their training times remain comparable.
Sparse MoE methods (MoLA and AdaMoLE) exhibit nearly double the training time of MoLoRA despite having the same parameter counts. This discrepancy occurs because, unlike traditional MoE approaches, PEFT-based MoE methods feature experts with significantly fewer parameters and computations relative to the frozen LLM backbone; besides, the sparse token-to-expert routing process within each batch brings additional computational overhead.
This phenomenon is also observable between Soft MoA and Sparse MoA. However, as the batch size increases (e.g., BS=8 in Figure~\ref{fig:time_memoery}(a)), two key effects emerge: (i) the sparse per-batch routing overhead becomes relatively less significant, and (ii) the reduction strategy of redundant expert computation in Sparse MoA becomes more effective, ultimately causing its training time to converge with Soft MoA's.

\begin{figure*}[!t]
  \includegraphics[width=0.495\linewidth]{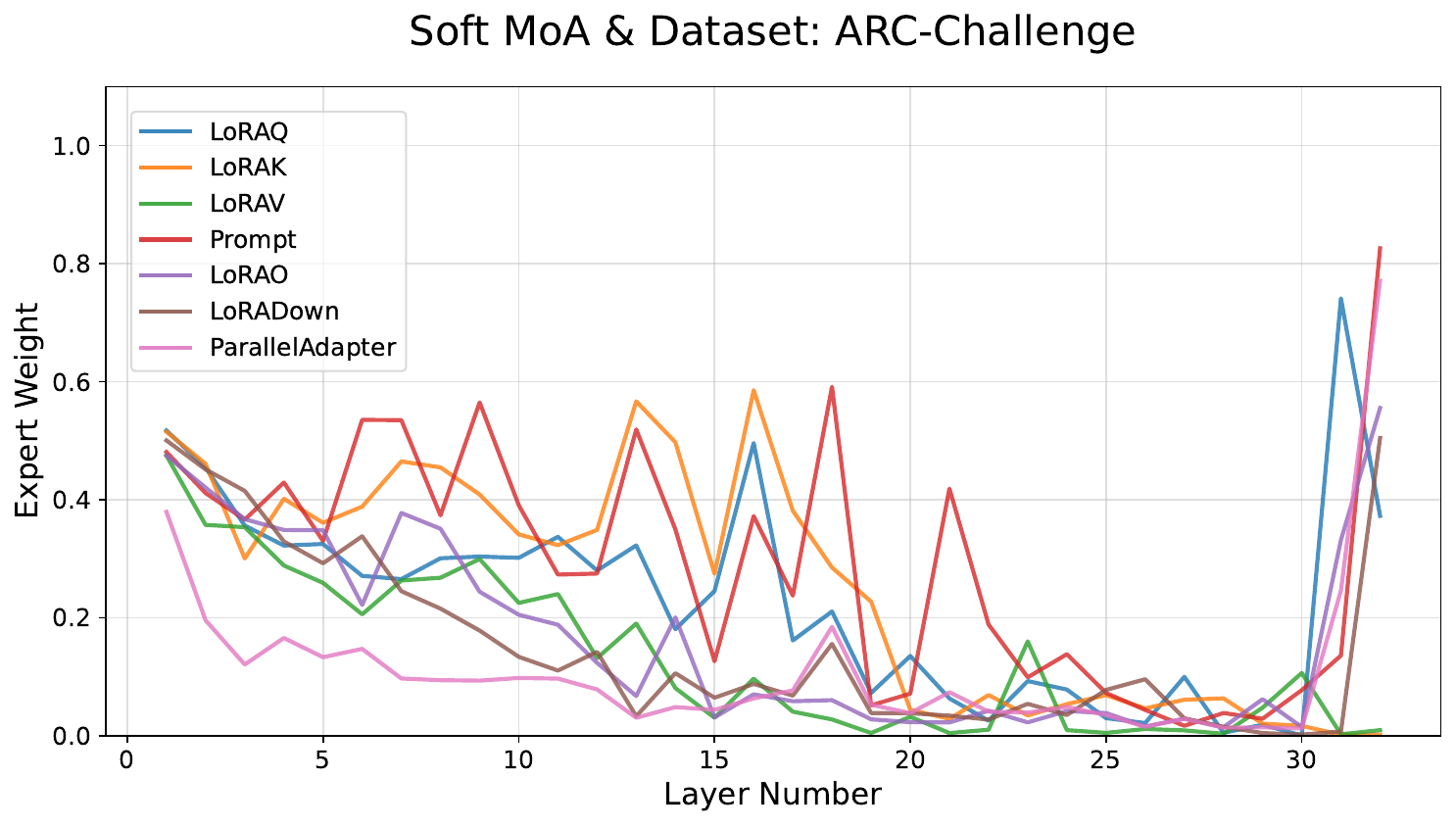} \hfill
  \includegraphics[width=0.495\linewidth]{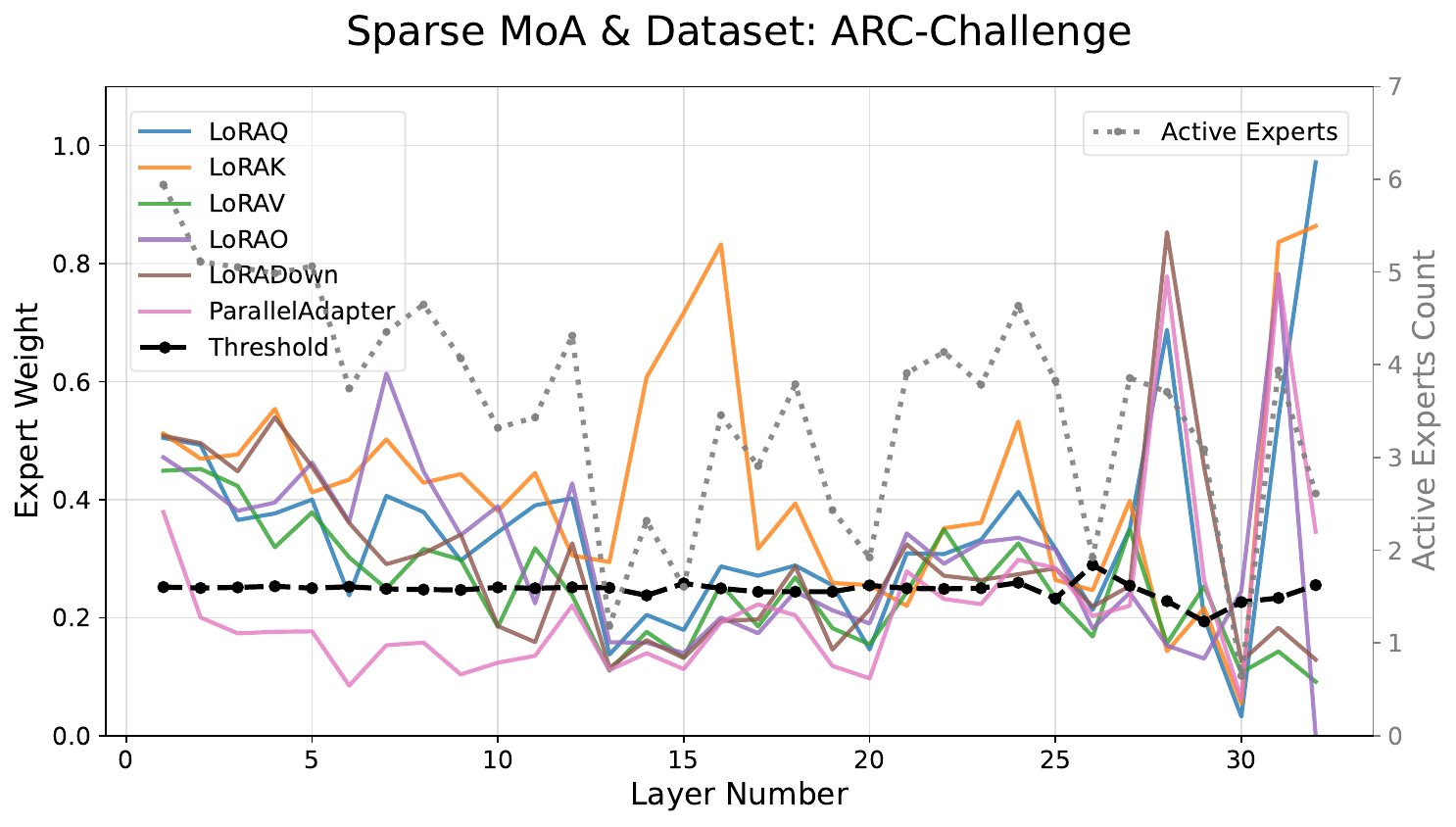}
  \caption {Visualization of average router weights per layer for Soft MoA (left) and Sparse MoA (right) on ARC-Challenge, averaged over tokens within 50 samples. Sparse MoA also includes the average per-layer threshold and the average count of activated experts. The average count of activated experts across layers in Sparse MoA is 3.55.}
  \label{fig:soft_vs_sparse_weights_ARCC}
\end{figure*}

As shown in Figure~\ref{fig:time_memoery}(b), both Soft MoA and Sparse MoA maintain memory requirements comparable to LoRA, with Sparse MoA demonstrating additional memory efficiency through its dynamic threshold function that selectively activates experts per token. This advantage becomes more pronounced at larger batch sizes - for instance, at BS=8, Sparse MoA reduces memory consumption by 3.86GB compared to Soft MoA.
Figure~\ref{fig:time_memoery}(c) reveals that the inference latency of both Soft MoA and Sparse MoA approaches that of standalone LoRA, while maintaining significant advantages over other MoE methods.

In summary, both MoA variants demonstrate notable advantages over MoE-LoRA methods across training time, GPU memory usage, and inference latency, nearing the performance of LoRA. Specifically:(i) Sparse MoA exhibits superior memory efficiency compared to Soft MoA through its sparse token-to-expert routing strategy.
(ii) The memory and computational benefits of Sparse MoA scale favorably with increasing batch size, making it particularly well-suited for resource-intensive applications.


      


\subsection{Difference between MoA and MoE}
\paragraph{The heterogeneous experts in the MoA method ensure expert specialization.}
As shown in Figure~\ref{fig:seeds_heatmap}, we compare the expert weight distributions across 32 Transformer layers on 50 test samples for Soft MoA and AdaMoLE, trained with different random seeds. Soft MoA exhibits highly consistent expert weight distributions across seeds, as illustrated in the first row. Experts in the lower 16 layers tend to be more active than those in the upper layers. Among the experts, LoRA(Q, K) and Prompt experts show higher activation than LoRA(O, V, Down) and the Parallel Adapter. This indicates that MoA is capable of leveraging the distinct adaptation capabilities of heterogeneous experts according to their structural characteristics and positions within the model.
In contrast, as shown in the second row, AdaMoLE fails to exhibit such consistency under different seeds, aligning with findings from prior work such as HMoE \citep{wang_hmoe_2024}, which identifies under-specialization as a common issue in homogeneous expert settings.

\paragraph{Inter-expert activation similarity.}
To analyze the architectural differences between MoA and homogeneous MoE, we measure expert specialization using the inter-expert activation standard deviation ($\sigma_{act}$). This metric is computed per token across all layers and subsequently averaged. A higher $\sigma_{act}$ indicates distinct expert specialization, whereas a lower value signifies "representation collapse"—a state where experts exhibit high redundancy and respond nearly identically to inputs. The metric is defined as:$$\sigma_{act} = \sqrt{\frac{1}{N}\sum_{i=1}^{N} (a_i - \bar{a})^2},$$where $N$ is the number of experts, $a_i$ is the activation of expert $E_i$, and $\bar{a}$ is the mean activation across all experts.

\begingroup
\begin{table}[ht]
  \centering
  \begin{tabular}{lccc}
    \Xhline{1.5pt}
    \rowcolor[HTML]{DAE0FB}
    \textbf{Model} & \textbf{Soft MoA} & \textbf{Sparse MoA} & \textbf{AdaMoLE}  \\ 
    \hline
    $\sigma_{act}$ ($\uparrow$) & 0.2330 & 0.2769 & 0.0643\\
   \Xhline{1.5pt}
  \end{tabular}
  \caption{\label{table:inter-expert}
    Comparison of Inter-expert Activation Standard Deviation between MoA and AdaMoLE on the AddSub Dataset.
  }
\end{table}

The results in the table~\ref{table:inter-expert} highlight a significant divergence between MoA and AdaMoLE. The minimal inter-expert activation standard deviation in AdaMoLE suggests a representation collapse, where experts fail to specialize. Conversely, MoA maintains a high degree of specialization, a benefit we attribute to its heterogeneous expert architecture.

\subsection{Expert Activation Comparison}

\paragraph{Sparse MoA significantly reduces expert computations with nearly no accuracy loss relative to Soft MoA.}
Figures~\ref{fig:soft_vs_sparse_weights_ARCC} and \ref{fig:soft_vs_sparse_weights_gsm8k} (see Appendix) present a comparison of expert activation weights between Soft MoA and Sparse MoA on ARC-Challenge and GSM8K datasets.
The distribution patterns of router weights for experts exhibit notable similarities between Soft MoA and Sparse MoA.
However, Soft MoA activates a fixed total number of experts per token, whereas Sparse MoA employs a thresholding function for selective expert activation. As illustrated in Figure 4 (right), the average number of activated experts in Sparse MoA drops to merely 1-2 in certain intermediate layers. The average number of activated experts across all layers is reduced from 6 to 3.55, reducing 40\% expert computations with a negligible accuracy drop of 0.3\%. \looseness=-1

\section{Conclusion}
We present MoA, a novel heterogeneous parameter-efficient fine-tuning (PEFT)  method that adapts LLMs to downstream tasks using PEFT adapters with diverse architectures and a minimal number of trainable parameters. 
Comprehensive experiments are conducted on commonsense and mathematical reasoning tasks, demonstrating that heterogeneous MoA outperforms the state-of-the-art homogeneous MoE-LoRA approaches, exhibiting superior performance in training time, inference latency, and GPU memory consumption. 


\newpage

\section*{Limitations}
While our proposed Sparse MoA method reduces GPU memory consumption compared to Soft MoA, its training and inference times are longer at a small batch size due to the additional computational overhead of sparse token-to-expert routing, achieving comparable or reduced time consumption only at larger batch sizes.
Furthermore, sparse routing methods, including Sparse MoA, necessitate token-to-expert assignment within a batch, thus limiting support to PEFT Adapters where input tokens within a sample can be computed independently. 
Consequently, methods like zero-initialized Prompt Tuning, where token computations within a sample are interdependent, are not compatible with Sparse MoA.



\bibliography{custom}
\clearpage

\appendix

\section{PEFT Methods And The Unified Form}
\label{sec:peft methods}

\paragraph{LoRA \citep{hu_lora_2021}:} LoRA injects trainable low-rank matrices into transformer layers to approximate the weight updates. For a pre-trained weight matrix $\boldsymbol{W} \in \mathbb{R}^{d \times k}$, LoRA represents its update with a low-rank decomposition $\boldsymbol{W} + \Delta W = \boldsymbol{W} + \boldsymbol{W}_{down}\boldsymbol{W}_{up}$, where $\boldsymbol{W}_{down} \in \mathbb{R}^{d \times r}$, $\boldsymbol{W}_{up} \in \mathbb{R}^{r \times k}$ are tunable parameters. For a specific input $\boldsymbol{x}$ to the linear projection in the transformer layer, the computation of LoRA is as follows:
\begin{equation}
    \boldsymbol{h} = \boldsymbol{xW} + \alpha \cdot \boldsymbol{xW}_{down}\boldsymbol{W}_{up},
\end{equation}
where $\alpha$ is a scalar hyperparameter.

\paragraph{Parallel Adapter \citep{he_towards_2022_paralleladapter}:} Parallel Adapters involve adding small adapter modules in parallel to the Feed-Forward Network (FFN) or Attention blocks within Transformer layers. The adapter module is normally moulded by a two-layer feed-forward neural network with a bottleneck: a down-projection with $\boldsymbol{W}_{down} \in \mathbb{R}^{d \times r}$ to project the input $\boldsymbol{x}$ to a lower-dimensional space specified by bottleneck dimension $r$, followed by a nonlinear activation function $f(\cdot)$, an up-projection with $\boldsymbol{W}_{up} \in \mathbb{R}^{r \times d}$ to project back to the input size. The parallel adapter can be defined as:
\begin{equation}
    \boldsymbol{h} = F(\boldsymbol{x}) + f(\boldsymbol{xW}_{down})\boldsymbol{W}_{up},
\end{equation}
where $F(\cdot)$ represents FFN or Attention block here.

\paragraph{Prompt tuning \citep{zhang_llama-adapter_2023}:} An advanced zero-initialized prompt tuning method adaptively incorporates instructional signals while preserving the pre-trained knowledge in LLMs. In each transformer layer, let $\boldsymbol{P} \in \mathbb{R}^{K \times d}$ denote the learnable prompts, the zero-initialized prompt tuning in the attention block computed as:
\begin{equation}
    \boldsymbol{Q} = \boldsymbol{xW}_q ,
\end{equation}
\begin{multline} \label{eq:prompt}
    \boldsymbol{h} = Attn(\boldsymbol{Q}, \boldsymbol{xW}_k, \boldsymbol{xW}_v) + \\ 
    g \cdot Attn(\boldsymbol{Q} , \boldsymbol{PW}_k, \boldsymbol{P}\boldsymbol{W}_{v}),
\end{multline}
where $g$ is a learnable scalar for each head initialized with zero, $\boldsymbol{W}_q , \boldsymbol{W}_k, \boldsymbol{W}_v$ are frozen pre-trained matix. Note that the first term in Equation~\ref{eq:prompt} is the original attention without prompts; only the second term includes trainable parameters.

\paragraph{The unified form:} As demonstrated above, LoRA, Parallel Adapter, and Prompt Tuning approaches can be simplified into a unified form:
\begin{equation}
    \boldsymbol{h} = F(\boldsymbol{x}) +  E(\boldsymbol{x}),
\end{equation}
where $F(\boldsymbol{x})$ denotes the transformer module to which the corresponding PEFT Adapter \(E\) is applied.

\section{Additional Experiments and Analyses}
\subsection{Effect of Threshold Function}

To validate the effect of the threshold function in Sparse MoA, we compare it against fixed-threshold baselines under varying hyperparameters \(\Gamma_{\text{max}} \in \{0.2, 0.5, 0.8\}\), with an additional setting of \(\Gamma_{\text{max}} = 1\) exclusive to the threshold function. As shown in Table~\ref{table:ablation_sparseMoA}, the threshold function achieves higher performance ceilings and consistently outperforms fixed-threshold methods across all \(\Gamma_{\text{max}}\) configurations. This demonstrates that tokens exhibit varying contextual importance: the dynamic threshold function assigns lower thresholds to critical tokens (promoting multi-expert collaboration) and higher thresholds to trivial ones (reducing redundant computations), thereby better leveraging the LLM’s capabilities while maintaining efficiency.  

\begingroup
\begin{table}[t]
  \centering
  \begin{tabular}{lcc}
    \Xhline{1.5pt}
    \rowcolor[HTML]{DAE0FB}
    \textbf{Max Gamma} & \textbf{Threshold} \textbf{Function} & \textbf{Math} \\
    \hline
    0.2 & \faClose & 81.22 \\
    0.5 & \faClose & 80.92 \\
    0.8 & \faClose & 78.58 \\
    0.2 & \faCheck & 81.24 \\
    0.5 & \faCheck & 81.51 \\
    0.8 & \faCheck & 81.36 \\
        \rowcolor[HTML]{E9F3FE}
    1 & \faCheck & 80.61 \\
   \Xhline{1.5pt}
  \end{tabular}
  \caption{\label{table:ablation_sparseMoA}
  Comparative performance of fixed-threshold baseline and threshold function in Sparse MoA under varying hyperparameters $\Gamma_{\text{max}}$.
  }
\end{table}

\begin{figure*}[ht]
  \includegraphics[width=0.495\linewidth]{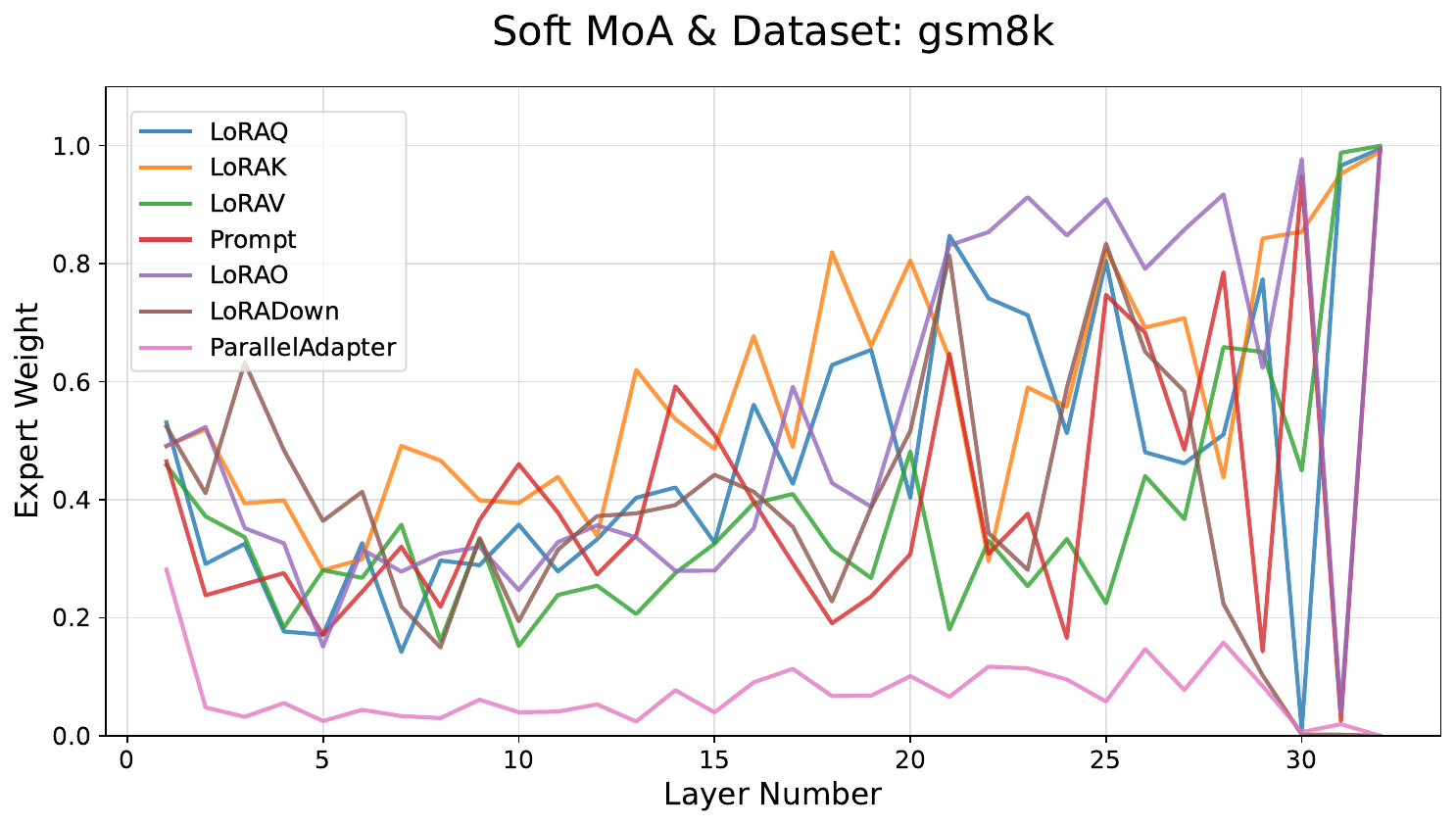} \hfill
  \includegraphics[width=0.495\linewidth]{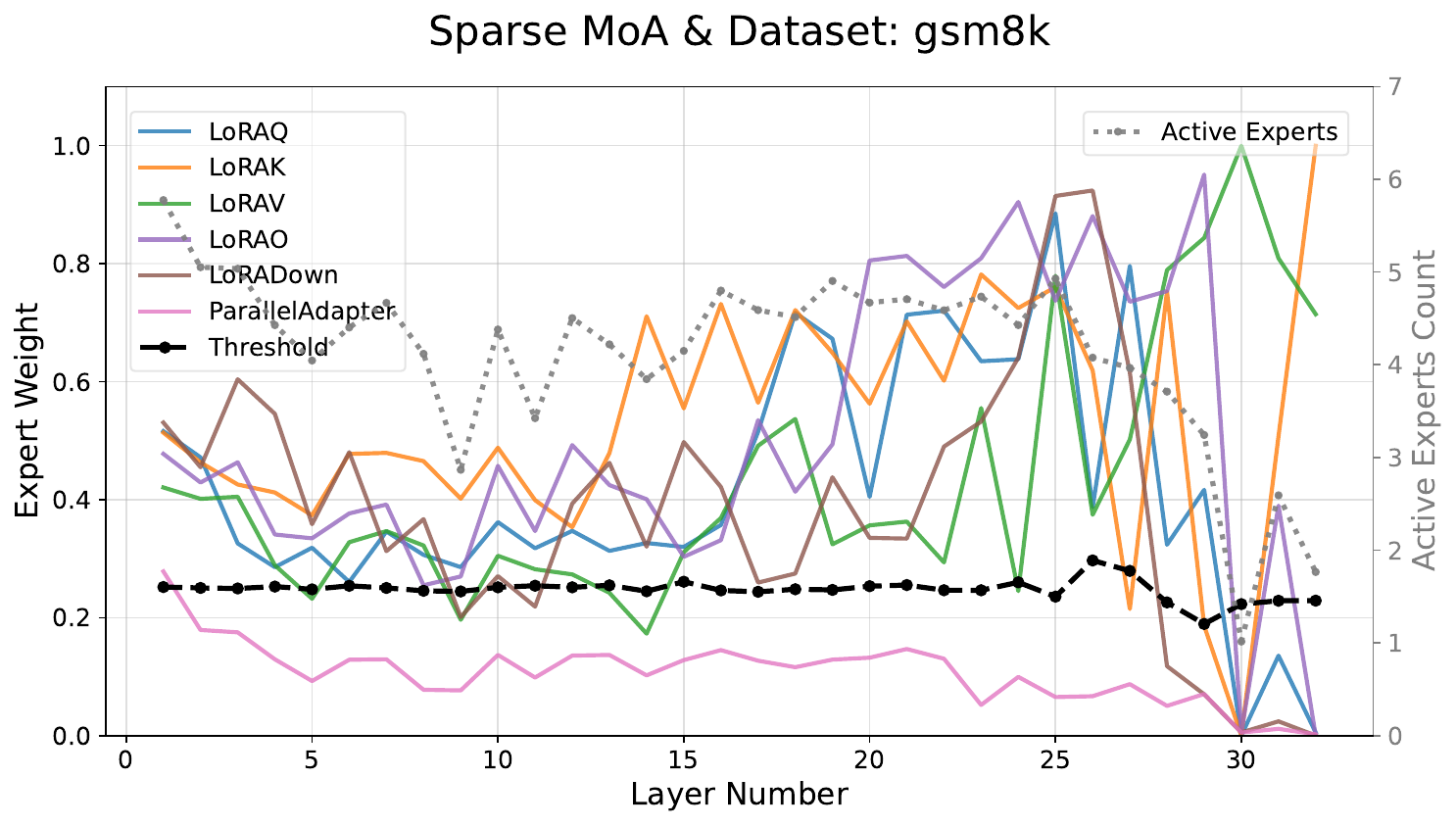}
  \caption {Visualization of average router weights per layer for Soft MoA (left) and Sparse MoA (right) on gsm8k, averaged over tokens within 50 samples. The Sparse MoA plot (right) also includes the average per-layer threshold and the average count of activated experts. The average count of activated experts across layers in Sparse MoA is 4.13.}
  \label{fig:soft_vs_sparse_weights_gsm8k}
\end{figure*}

\subsection{Comparison between Instance-level and Token-level Routing Methods} \label{sec:instance-token-level}
In contrast to token-level routing approaches, instance-level routing, wherein the model makes a single routing decision for the entire input example, significantly reduces the computational overhead associated with the router. This brings the overall computational cost closer to that of standard, non-MoE architectures. We conducted a comparison between token-level MoA and the instance-level variant of Soft MoA and the UniPEFT approach on LLaMA3.1-8B.
UniPEFT utilizes a separate gated network for each type of adapter to control their influence on the input sample. As presented in Table~\ref{table:instance level}, the results indicate that both instance-level approaches yield performance inferior to token-level routing methods. Furthermore, their performance is also surpassed by employing a standard LoRA adaptation alone.

This performance deficit can be attributed to key limitations of instance-level routing in PEFT of LLM. Firstly, these methods typically rely on the mean pooling of input text hidden states as the router's input, which fails to capture fine-grained characteristics of the sample adequately. Secondly, the constraint of making a single routing choice for the entire sequence precludes the possibility of dynamically adjusting the contributions of different experts in response to the diverse contextual nuances within the input.
\begingroup
\setlength{\tabcolsep}{2pt}
\begin{table}[ht]
\small
  \centering
  \begin{tabular}{lccc}
    \Xhline{1.5pt}
    \rowcolor[HTML]{DAE0FB}
    \textbf{Model} & \textbf{Routing level} & \textbf{Commonsense15K} & \textbf{Math14K}  \\ 
    \hline
    UniPEFT & instance & 78.29 & 74.65 \\
    Soft MoA & instance & 73.18 & 76.11 \\
    Soft MoA & token  & 84.96 &  81.51 \\
    Sparse MoA & token & 84.62 & 81.20\\
   \Xhline{1.5pt}
  \end{tabular}
  \caption{\label{table:instance level}
  Comparison between the token-level MoA approaches and instance-level methods. The instance-level methods include an instance-level variant of Soft MoA and the UniPEFT method.
  }
\end{table}

\subsection{Foundation Model Analysis} \label{appendix:qwen3_results}
To validate the effectiveness of our method across different and larger foundation models, we conducted experiments on Qwen3-8B and the larger-scale Qwen3-14B \cite{yang2025qwen3}. For the experiments on Qwen3-8B, we adopted the same experimental settings as LLaMA-3.1 8B, as described in Section~\ref{sec:setting}. However, when using Qwen3-14B, we encountered GPU Out-of-Memory (OOM) issues for MoLoRA, HydraLoRA, and AdaMoLE when the number of experts was set to 8. Consequently, for these three methods, we set the number of experts to 4. In contrast, for MoLA, we still maintained a total of 8 experts, and only activated the top 2.

As detailed in Tables~\ref{table:math14k_qwen3-8b} and ~\ref{table:math14k_qwen3-14b}, our proposed Soft MoA and Sparse MoA methods outperform both LoRA and other MoE-LoRA approaches on mathematical reasoning benchmarks.
Specifically, on the Qwen3-8B model, Soft MoA surpasses competing methods that utilize 2 to 4 times more parameters. On the larger Qwen3-14B model, Sparse MoA achieves the highest performance among all baselines while using the fewest parameters (34.50M).
These results collectively demonstrate the superior parameter efficiency of the heterogeneous MoA approach compared to conventional MoE-LoRA methods, and further affirm its robustness across different foundation models and scales.
\begingroup
\setlength{\tabcolsep}{4pt}
\begin{table*}[t]
\small
  \centering 
  \begin{tabular}{lcccccccc}
    \Xhline{1.5pt}
    \rowcolor[HTML]{DAE0FB}
    \textbf{Models} & \textbf{AddSub} & \textbf{AQuA} & \textbf{GSM8k} & \textbf{MultiArith} & \textbf{SingleEq} & \textbf{SVAMP} & \textbf{Average} & \textbf{Param} \\
    \hline
    \hline
    \rowcolor[HTML]{D3D3D3}
    LoRA & $93.67_{\pm 0.39}$ & $37.40_{\pm 2.58}$ & $81.80_{\pm 0.67}$ & $98.17_{\pm 0.86}$ & $96.85_{\pm 0.23}$ & $88.80_{\pm 0.53}$ & $82.78_{\pm 0.47}$ & 24.77M \\ 
    MoLoRA & $93.76_{\pm 1.30}$ & $42.26_{\pm 2.41}$ & $82.51_{\pm 2.35}$ & $98.11_{\pm 0.38}$ & $96.06_{\pm 1.04}$ & $90.40_{\pm 0.30}$ & $\underline{83.85}_{\pm 1.17}$ &  107.35M \\
    HydraLoRA & $93.16_{\pm 1.27}$ & $39.11_{\pm 2.79}$ & $81.78_{\pm 1.48}$ & $98.11_{\pm 0.25}$ & $95.21_{\pm 0.75}$ & $88.80_{\pm 0.26}$ & $82.70_{\pm 0.88}$ & 49.55M\\
    MoLA & $93.33_{\pm 0.89}$ & $42.78_{\pm 0.45}$ & $81.50_{\pm 1.14}$ & $98.33_{\pm 0.44}$ & $96.13_{\pm 0.50}$ & $89.83_{\pm 1.16}$ & $83.65_{\pm 0.26}$ & 107.35M \\
    AdaMoLE & $92.24_{\pm 1.05}$ & $41.73_{\pm 4.26}$ & $81.65_{\pm 1.00}$ & $98.22_{\pm 0.25}$ & $96.19_{\pm 0.80}$ & $89.10_{\pm 0.75}$ & $83.19_{\pm 0.85}$ & 108.38M \\ 
    FlyLoRA & $93.00_{\pm 0.81}$ & $39.11_{\pm 1.98}$ & $86.20_{\pm 0.50}$ & $98.44_{\pm 0.10}$ & $95.87_{\pm 0.86}$ & $89.23_{\pm 0.70}$ & $83.64_{\pm 0.40}$ & 24.77M \\
    \hline
    \rowcolor[HTML]{E9F3FE}
    Soft MoA & $93.50_{\pm 0.15}$ & $43.31_{\pm 4.09}$ & $85.17_{\pm 1.03}$ & $98.06_{\pm 0.51}$ & $96.72_{\pm 0.50}$ & $88.30_{\pm 0.72}$ & $\boldsymbol{84.17}_{\pm 0.89}$ & 26.99M \\  
    \rowcolor[HTML]{E9F3FE}
    Sparse MoA &  $93.33_{\pm 0.15}$ & $40.68_{\pm 2.95}$ & $83.85_{\pm 0.35}$ & $98.17_{\pm 0.44}$ & $97.31_{\pm 0.11}$ & $88.80_{\pm 1.66}$ & $83.69_{\pm 0.76}$ & 24.49M \\ 
   \Xhline{1.5pt}
  \end{tabular}
  \caption{\label{table:math14k_qwen3-8b}
  Experimental results on Qwen3-8B for mathematical reasoning benchmarks. The evaluation metric is accuracy. All methods are run with three random seeds; mean and standard deviation are reported. ``Param'' indicates trainable parameters.
  }
\end{table*}

\begingroup
\setlength{\tabcolsep}{4pt}
\begin{table*}[t]
\small
  \centering
  \begin{tabular}{lcccccccc}
    \Xhline{1.5pt}
    \rowcolor[HTML]{DAE0FB}
    \textbf{Models} & \textbf{AddSub} & \textbf{AQuA} & \textbf{GSM8k} & \textbf{MultiArith} & \textbf{SingleEq} & \textbf{SVAMP} & \textbf{Average} & \textbf{Param} \\
    \hline
    \hline
    \rowcolor[HTML]{D3D3D3}
    LoRA & $96.20_{\pm 0.25}$ & $41.34_{\pm 3.39}$ & $87.49_{\pm 0.86}$ & $97.83_{\pm 0.29}$ & $97.64_{\pm 0.41}$ & $90.60_{\pm 0.56}$ & $85.18_{\pm 0.86}$ & 35.39M \\ 
    MoLoRA & $96.37_{\pm 0.29}$ & $44.88_{\pm 2.73}$ & $87.59_{\pm 1.12}$ & $98.33_{\pm 0.29}$ & $97.90_{\pm 0.45}$ & $91.50_{\pm 0.69}$ & $86.10_{\pm 0.72}$ &  76.84M \\ 
    HydraLoRA & $95.78_{\pm 1.39}$ & $43.70_{\pm 2.39}$ & $88.05_{\pm 1.20}$ & $98.17_{\pm 1.04}$ & $97.77_{\pm 0.41}$ & $91.37_{\pm 0.81}$ & $85.81_{\pm 0.34}$ & 40.47M\\ 
    MoLA & $96.96_{\pm 0.67}$ & $42.26_{\pm 4.55}$ & $88.98_{\pm 0.68}$ & $98.67_{\pm 0.17}$ & $97.83_{\pm 0.52}$ & $91.63_{\pm 0.57}$ & $86.06_{\pm 0.66}$ & 153.68M \\
    AdaMoLE & $96.20_{\pm 0.25}$ & $40.55_{\pm 4.54}$ & $88.02_{\pm 1.26}$ & $98.50_{\pm 0.17}$ & $98.10_{\pm 0.23}$ & $91.13_{\pm 0.70}$ & $85.42_{\pm 0.54}$ & 78.36M \\ 
    FlyLoRA & $95.27_{\pm 0.15}$ & $41.86_{\pm 2.24}$ & $88.53_{\pm 0.52}$ & $98.11_{\pm 0.35}$ & $96.72_{\pm 0.50}$ & $91.47_{\pm 0.15}$ & $85.33_{\pm 0.40}$ & 35.39M \\
    \hline
    \rowcolor[HTML]{E9F3FE}
    Soft MoA & $96.96_{\pm 0.25}$ & $44.88_{\pm 2.98}$ & $88.78_{\pm 0.54}$ & $98.33_{\pm 0.10}$ & $98.62_{\pm 0.80}$ & $90.80_{\pm 0.25}$ & $\boldsymbol{86.40}_{\pm 0.72}$ & 37.98M \\ 
    \rowcolor[HTML]{E9F3FE}
    Sparse MoA &  $96.29_{\pm 0.89}$ & $44.62_{\pm 3.44}$ & $88.20_{\pm 1.02}$ & $98.11_{\pm 0.35}$ & $98.10_{\pm 0.11}$ & $91.63_{\pm 0.99}$ & $\underline{86.16}_{\pm 0.77}$ &  34.50M \\ 
   \Xhline{1.5pt}
  \end{tabular}
  \caption{\label{table:math14k_qwen3-14b}
  Experimental results on Qwen3-14B for mathematical reasoning benchmarks. The evaluation metric is accuracy. All methods are run with three random seeds; mean and standard deviation are reported. ``Param'' indicates trainable parameters.
  }
\end{table*}

\subsection{Code generation tasks} \label{sec:code_task}
\paragraph{Datasets and Evaluation}
To evaluate the code generation performance of our method, we utilize the CodeAlpaca-20k \cite{chaudhary2023codealpaca} dataset as our training set. The evaluation is conducted on the HumanEval \cite{chen2021humaneval} benchmark. We use the pass@$k$ metrics (where $k \in \{1, 5, 10\}$) to measure the performance on code generation tasks. The pass@$k$ denotes the probability of generating at least one functionally correct solution within $k$ attempts.

\paragraph{Experimental setting}
For the code generation task, we employ Qwen3-8B and Llama3.1-8B as base models. We adopt hyperparameter configurations similar to those used for mathematical reasoning. Given that different baselines exhibit sensitivity to the number of training steps on the code generation task, we perform checkpoint selection using the validation set during training. Specifically, we train for 5 epochs on Qwen3-8B and 1 epoch on Llama3.1-8B. For code generation, the maxmium sequence length is set to 500.

\paragraph{Main results}
As illustrated in Table~\ref{table:code_result}, our method outperforms both MoE-LoRA and LoRA-based baselines (e.g., DenseLoRA and FlyLoRA) while utilizing the fewest parameters. Specifically, on Qwen3-8B, Sparse MoA surpasses LoRA by 10.61 points in pass@1 and exceeds the strongest baseline, HydraLoRA, by 3.41 points. We observe that sparse MoE-LoRA methods, such as MoLA and AdaMoLE, exhibit suboptimal performance on code generation tasks compared to soft-weighted variants like MoLoRA and HydraLoRA. Furthermore, while the performance of FlyLoRA fluctuates significantly between Llama 3.1-8B and Qwen3-8B, our MoA method maintains consistent superiority across both base models, demonstrating its robust generalizability.
\begingroup
\setlength{\tabcolsep}{2pt}
\begin{table*}[ht]
  \centering
  \begin{tabular}{lcccrcccr}
    \Xhline{1.5pt}
    \rowcolor[HTML]{DAE0FB}
    & \multicolumn{4}{|c}{\texttt{Llama3.1-8B}} & \multicolumn{4}{|c}{\texttt{Qwen3-8B}} \\
    \hline
    \hline
    \rowcolor[HTML]{DAE0FB}
    \textbf{Methods}  & \textbf{Pass@1} & \textbf{Pass@5} & \textbf{Pass@10} &  \textbf{\# Parameters} & \textbf{Pass@1} & \textbf{Pass@5} & \textbf{Pass@10} &  \textbf{\# Parameters} \\
    \hline
    LoRA & 26.22 & 54.41 & 67.07 & 23.06M & 39.69 & 72.29 & 81.71 & 24.77M\\
    MoLoRA & \underline{34.15} & 63.15 & \textbf{72.56} & 100.14M & 43.78 & 73.49 & 82.93 & 107.35M\\
    HydraLora & 33.35 & 60.96 & 69.51 & 45.09M & 46.89 & 75.61 & 84.15 & 49.55M\\
    MoLA & 28.72 & 56.53 & 67.68 & 100.14M & 41.10 & 71.60 & 81.10 & 107.35M \\
    AdaMoLE & 21.52 & 44.58 & 56.10 & 101.12M & 40.91 & 66.65& 76.83 & 108.38M \\
    DenseLoRA & 31.58 & 58.71 & 67.07 & 26.74M & 46.82 & 76.04 & 84.75 & 27.00M \\
    FlyLoRA & 32.32 & \underline{63.21}	& \underline{71.95} & 23.06M & 20.18 & 38.15 & 48.17 & 24.77M \\
    \hline
    \rowcolor[HTML]{E9F3FE}
    SoftMoA & 31.10 & 59.86 & 70.73 & 24.52M & \textbf{50.61} & \underline{80.57} & \textbf{89.02} & 26.70M \\
    \rowcolor[HTML]{E9F3FE}
    SparseMoA & \textbf{35.55} & \textbf{63.26} & 70.73 & 22.29M & \underline{50.30} & \textbf{81.12} & \underline{87.81} & 24.49M \\
    \Xhline{1.5pt}
  \end{tabular}
  \caption{\label{table:code_result}
    Experimental results for the code generation task on Llama3.1-8B and Qwen3-8B. The evaluation metric is pass@k. "\# Parameters" indicates trainable parameters. Bold numbers indicate the highest performance scores and underline numbers indicate the second performance scores.
  }
\end{table*}
\endgroup

\subsection{Case Study of Router Weight} \label{sec:case study}
Figures~\ref{fig:softmoa_case} and \ref{fig:sparsemoa_case} illustrate the router weights for a specific sample in a particular layer for Soft MoA and Sparse MoA models, respectively.
\begin{figure*}[t]
  \includegraphics[width=1\textwidth, height=24cm]{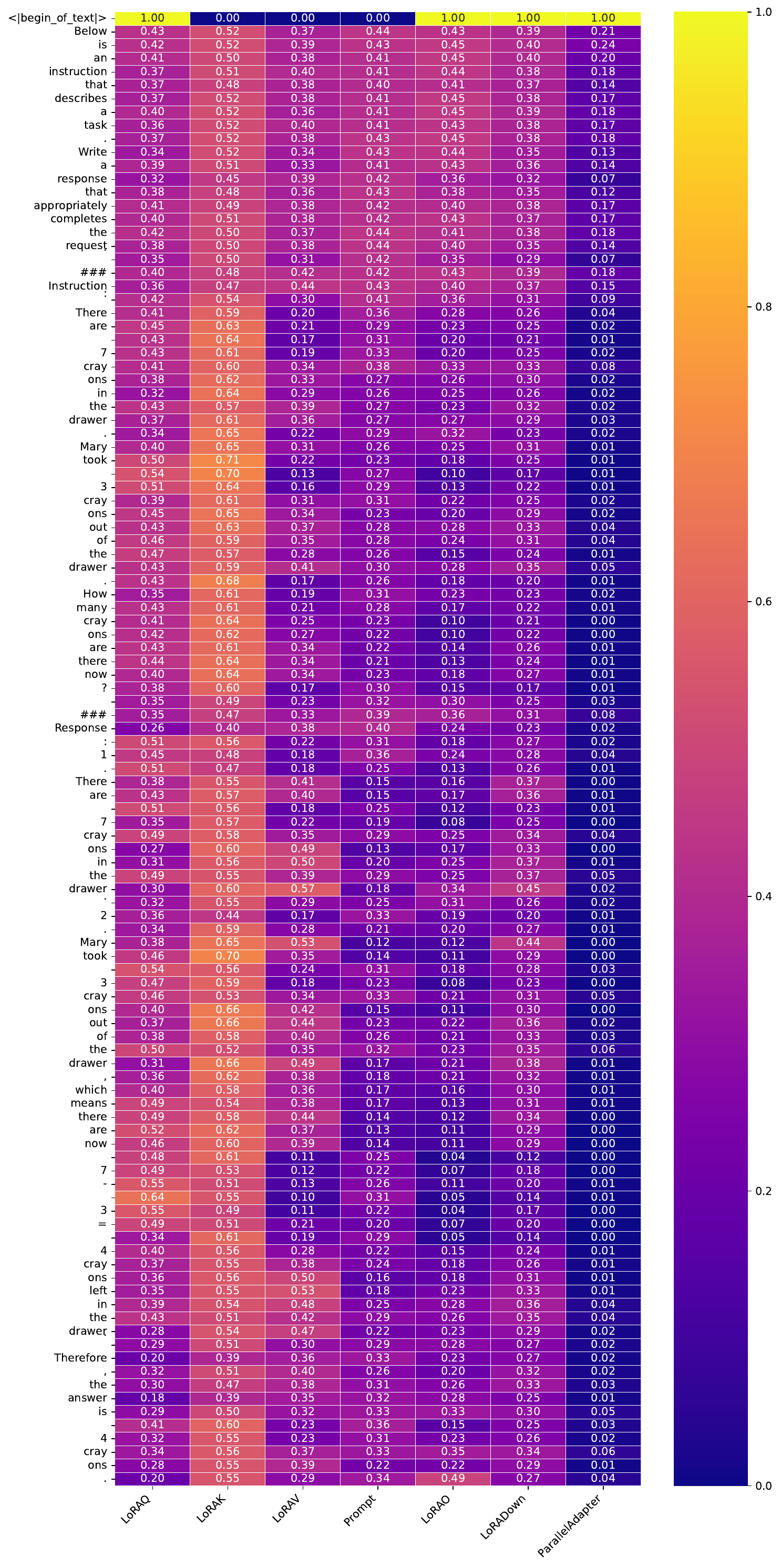}
  \caption{Router Weights of an example at Layer 14 in Soft MoA.}
  \label{fig:softmoa_case}
\end{figure*}

\begin{figure*}[t]
  \includegraphics[width=1\textwidth, height=24cm]{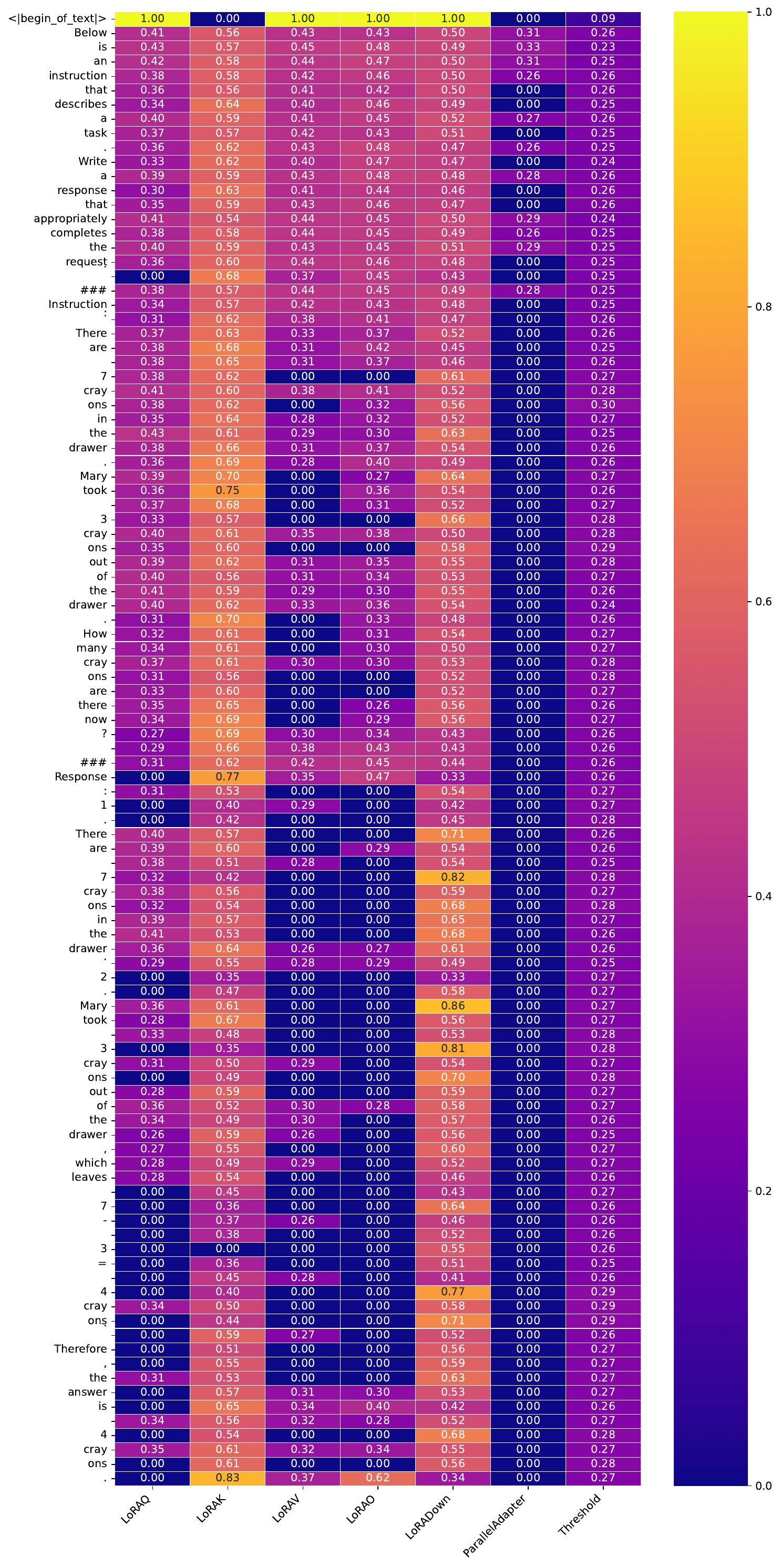}
  \caption{Router Weights of an example at Layer 14 in Sparse MoA. "Threshold" denotes the dynamic value of the threshold function for Sparse MoA. A router weight of "0.00" for an expert indicates that the expert's computation is skipped. Sparse MoA avoids unnecessary computations from the low-contribution expert for each token.}
  \label{fig:sparsemoa_case}
\end{figure*}




%
\begin{figure}[t]
  \includegraphics[width=\columnwidth]{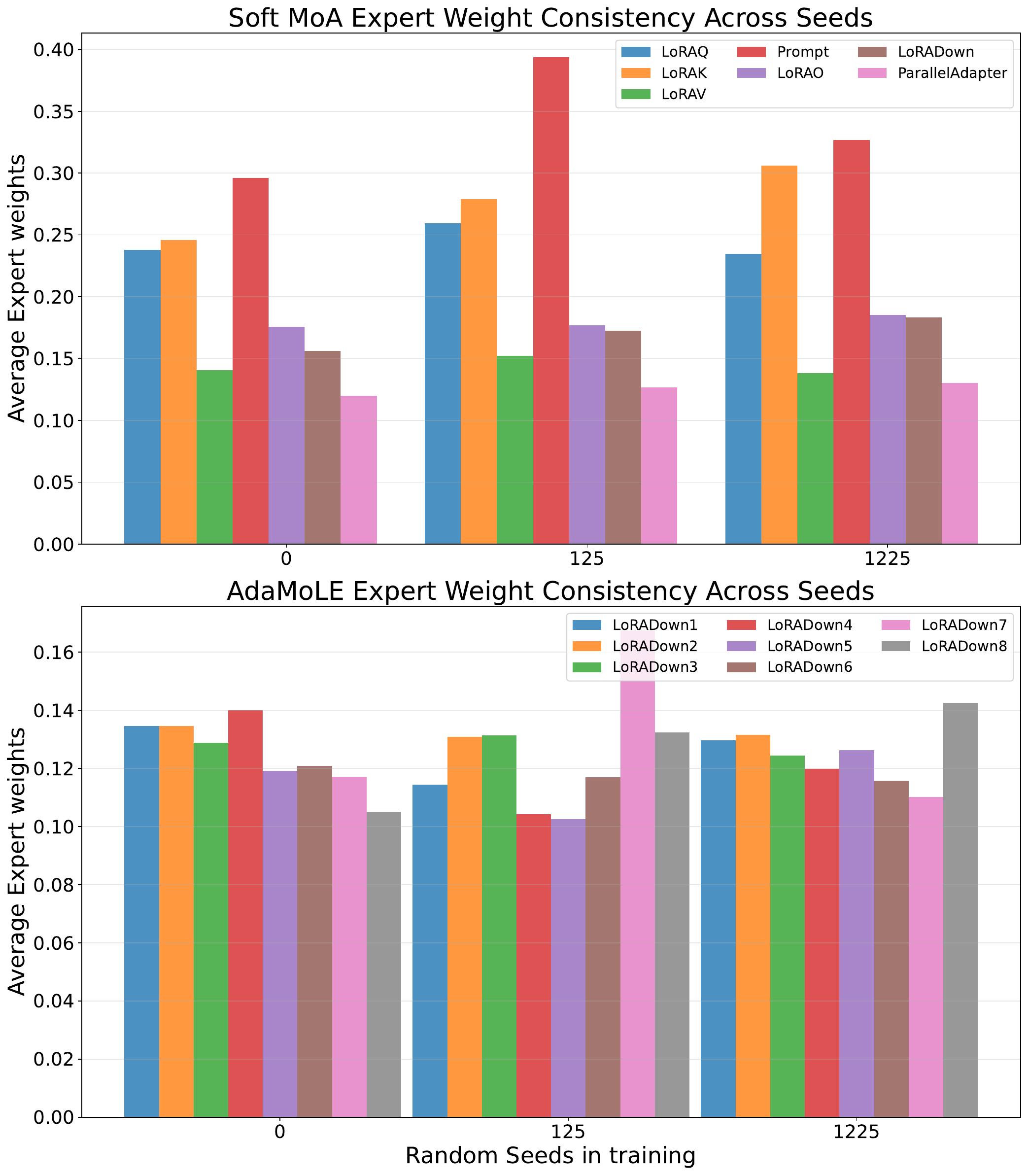}
  \caption{Average router weights across layers. Soft MoA exhibits strong consistency under different random seeds in training, whereas AdaMoLE does not.}
  \label{fig:seeds_average}
\end{figure}


\end{document}